\documentclass[sigconf]{acmart}
\AtBeginDocument{%
  }

\copyrightyear{2024} 
\acmYear{2024} 
\setcopyright{rightsretained} 
\acmConference[SETN 2024]{13th Conference on Artificial Intelligence}{September 11--13, 2024}{Piraeus, Greece}
\acmBooktitle{13th Conference on Artificial Intelligence (SETN 2024), September 11--13, 2024, Piraeus, Greece}\acmDOI{10.1145/3688671.3688775}
\acmISBN{979-8-4007-0982-1/24/09}

\usepackage{algorithm}
\usepackage{algpseudocode}
\usepackage{balance}
\begin{document}

\title{Local Explanations and Self-Explanations for Assessing Faithfulness in black-box LLMs}
\author{Christos Fragkathoulas}
\authornote{Both authors contributed equally to this research.}
\orcid{0009-0002-1697-1912}
\affiliation{%
  \institution{University of Ioannina and Archimedes/Athena RC}
  \city{Athens}
  \country{Greece}
}
\email{ch.fragkathoulas@athenarc.gr}

\author{Odysseas S. Chlapanis}
\authornotemark[1]
\affiliation{%
  \institution{Athens University of Economics and Business and Archimedes/Athena RC}
  \city{Athens}
  \country{Greece}}
\email{odyhlapanis@aueb.gr}

\begin{abstract}
This paper introduces a novel task to assess the faithfulness of large language models (LLMs) using local perturbations and self-explanations. Many LLMs often require additional context to answer certain questions correctly. For this purpose, we propose a new efficient alternative explainability technique, inspired by the commonly used leave-one-out approach. Using this approach, we identify the sufficient and necessary parts for the LLM to generate correct answers, serving as explanations. We propose a metric for assessing faithfulness that compares these crucial parts with the self-explanations of the model. Using the Natural Questions dataset, we validate our approach, demonstrating its effectiveness in explaining model decisions and assessing faithfulness.
\end{abstract}



\begin{teaserfigure}
  \includegraphics[width=\textwidth]{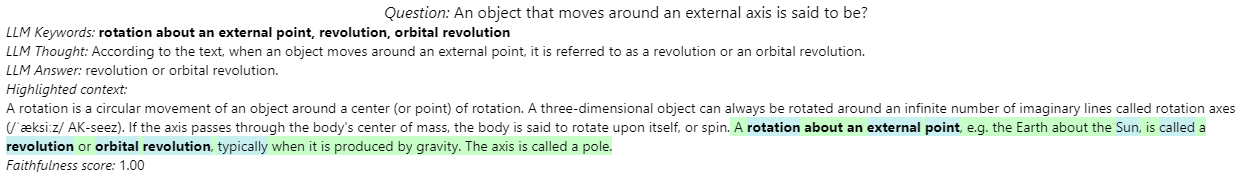}
  %\vspace{-2.2em}
  \caption{An example of our proposed black-box explainability algorithm (in color) and self-explanations (in bold). 
  % The question and the context are given as input to the LLM.
  }
  \Description{}
  \label{fig: example_thought}
\end{teaserfigure}


\maketitle

% ACM Reference Format:
% Christos Fragkathoulas and Odysseas S. Chlapannis. 2024. Local Explanations and Self-Explanations for Assessing Faithfulness in black-box LLMs. In *Proceedings of the 1st Workshop on Responsible Artificial Intelligence (ReAI ’24)*, September 11–13, 2024, Piraeus, Athens, Greece. ACM, New York, NY, USA, 5 pages. https://doi.org/10.1145/3688671.3688775

\section{Introduction}
Artificial intelligence (AI) and machine learning models have become ubiquitous in various domains, ranging from healthcare to finance and beyond. 
Large language models (LLMs) have demonstrated remarkable capabilities in understanding and generating human-like text. However, 
many top-performing models are proprietary and accessible only via APIs, acting as black boxes and offering little insight into their thought processes. 
This opacity poses significant challenges \cite{weidinger2021ethical}, especially in applications where understanding the decision-making process of a model is critical. LLMs are also known to suffer from issues such as hallucinations \cite{zhang2023siren}, where they generate plausible-sounding but incorrect or nonsensical answers and the generation of overly verbose outputs that may obscure the relevant information \cite{zhao2024explainability}.
Even when models provide a chain-of-thought explanation \cite{wei2022chain}, it may not reflect their actual reasoning \cite{parcalabescu2023measuring}. This underscores the urgent need for faithful explainability in AI.

Explainability refers to the capability of explaining or describing the behavior of models in terms that are comprehensive to humans \cite{dozat2016deep, du2019techniques}. 
Explainable AI is crucial in helping users trust and effectively utilize AI systems \cite{choung2023trust, liu2023increasing}, enabling developers to debug and improve models \cite{bastings-etal-2022-will, strobelt2018s, yuksekgonul2022post}, ensuring compliance with regulatory standards 
by providing transparency into automated decision-making processes
\cite{9060413},
to help in bias identification, provide causal understanding, and mitigate biases within models \cite{fragkathoulas2024explaining}.

Both surveys \cite{choudhary2022interpretation, zhao2024explainability} review extensive research on language model explainability, including attention mechanisms, gradient-based explanations, and post-hoc techniques.
% Both surveys \cite{choudhary2022interpretation, zhao2024explainability} highlight the extensive research conducted in explainability of language models, covering a variety of methods ranging from attention mechanisms to gradient-based explanations and post-hoc interpretation techniques.
% Gradient-based methods link input features to outputs for better interpretability, while attention-based mechanisms focus on the most relevant parts of the input data, providing insights into the decision-making process \cite{choudhary2022interpretation}.
Many works leverage gradients to directly link input features to model outputs, enhancing the interpretability of NLP models, while others utilize attention-based mechanisms to focus on the most relevant parts of the input data, providing insights into the decision-making process \cite{choudhary2022interpretation}. 
However, these methods are not applicable to commercially available LLMs where internal architectures are inaccessible.
Other approaches employ perturbations on the input to observe changes in model behavior, such as Shapley-based approaches \cite{shapley1953value}, which involve numerous attempts to determine the impact of each feature \cite{mosca2022shap}.
Our approach lies in the field of local post-hoc black-box techniques to study proprietary LLMs with text-only access.
It offers a more efficient alternative, providing useful insights into model behavior with a constant number of API calls per sample. Specifically, we use a novel algorithm inspired by the commonly used leave-one-out (LOO) approach to identify the crucial parts of the context that LLMs rely on to generate correct answers. By systematically omitting parts of the provided context, we can determine specific important words that without them the model is unable to respond correctly. 
We assess the faithfulness of the model by comparing these findings with the model’s own explanations. We conduct our study using a real QA dataset, which contains real user-generated questions requiring models to leverage snippets from Wikipedia articles to respond correctly.

The remainder of this paper is structured as follows.
In Section \ref{seq: problem}, we introduce the problem, in Section \ref{seq: method} our methodology, and in Section \ref{seq: experiments}, experiments. In Section \ref{seq: relatedwork}, we discuss related work, and in Section \ref{seq: conclusion_futurework}, we offer conclusions.

\section{Problem Definition}
\label{seq: problem}
We introduce a novel task to assess the faithfulness of the large language models (LLMs) self-explanations on question-answer (QA) benchmarks that provide a helpful context snippet. 
Given question, context and answer triplets \(\mathcal{D} = \{(q_i, c_i, e_i)\}_{i=1}^N\), where \(q_i\) is a question, \(c_i\) is the corresponding context, and \(e_i\) is the ground truth answer, the goal is to assess how faithfully LLMs generate self-explanations (in the form of keywords taken from the context \(c_i\)) that align with their usefulness in generating correct answers. This involves identifying the crucial parts of the context that the model relies on to generate the answer.

Let \(M\) be the language model. For each question-context pair \((q_i, c_i)\), the model generates a response in the form of a triplet: thought \(t_i\), keywords \(k_i\), and answer \(a_i\). For simplicity, we ignore the thought and the keywords which are only used for visualization, and denote the response of the model as $a_i$:

%\vspace{-1.2em}
\[
a_i = M(q_i, c_i)
\]
%\vspace{-1.6em}

Our objective is twofold:

\begin{itemize}
    \item Identify the \textit{sufficient regions} \(SR_i\) set within the context $c_i$ that contain sufficient information for the model to answer correctly. Formally, let \(s \in SR_i\), \(s \subseteq c_i\), iff \(M(q_i, s) = e_i\). 
    
    \item Within a sufficient region \(s \in SR_i\), pinpoint the necessary keywords \(NK_s\) whose masking results in the model providing an incorrect answer. Formally, let \(t \in NK_s\), where \(t \subseteq s\), iff \(M(q_i, s.mask(t)) \neq e_i\). We define mask as the function that replaces the string t with the underscore `\_' in a string s.
\end{itemize}

\section{Method}
\label{seq: method}

\subsection{Dataset}
\label{seq: dataset}
For this analysis, we use the Natural Questions dataset \cite{kwiatkowski2019natural} which is designed to spur the development of open-domain question-answering systems and it has been used for benchmarking at QA studies like \cite{guu2020retrieval, chen2017reading, lee2019latent, yang2019end, lewis2020retrieval, liu2024lost}. Specifically, we are using the same context snippets as in \cite{liu2024lost}.
It contains questions from real users and requires systems to read and comprehend segments of Wikipedia articles to find answers.
A QA example of this dataset is:
\begin{itemize}
    \item \textbf{Question:} When did the watts riot start and end?
    \item \textbf{Long Answer:} The Watts riots, sometimes referred to as the Watts Rebellion, took place in the Watts neighborhood of Los Angeles from August 11 to 16, 1965.
    \item \textbf{Short Answer:} August 11 to 16 , 1965
\end{itemize}
In figure \ref{fig: example_thought} the long answer is the highlighted context (without the highlights) and the short answer is the same as the LLM Answer.

\subsection{Retrieval-Hard subset}
\label{seq: retrievalhardsubset}
Evaluating black-box models is challenging due to their unknown pretraining corpus. Models might rely on internal knowledge rather than context, leading to unfair comparisons. To address this, we use the \textit{Retrieval-Hard subset}, which includes only those samples the model fails to answer correctly without context. This framework is applicable to any dataset that has retrieved helpful context snippets and is useful for evaluating LLMs fairly in this setup.

\subsection{QA Evaluation}
Given a short answer $a_i$ from the model $M$ for a question $q_i$ and a specified context $c_i$, our goal is to evaluate the correctness of the short answer of the model.
Unlike previous work \cite{guu2020retrieval, chen2017reading}, who are using only the exact-match accuracy metric for evaluation, we have designed a hybrid metric that combines the results of exact-match, normalized exact-match, fuzzy exact-match, model-based embedding cosine similarity and date transformations. 
Exact match often fails due to natural language variability, such as different formatting of names or dates. Our hybrid metric addresses these issues by capturing semantic equivalence and format variations, offering a more robust evaluation of the answers of the model.
The mathematical formulation of these metrics is as follows:
\begin{flalign*}
    &\text{ExactMatch}(e_i, a_i) = \![ e_i = a_i ]\! &\\
    &\text{NormExactMatch}(e_i, a_i) = \text{ExactMatch}(norm(e_i), norm(a_i)) &\\
    &\text{FuzzyExactMatch}(e_i, a_i) = \![fuzzyMatch(e_i,a_i) \geq 90]\! &\\
    &\text{EmbedSimilarity}(e_i,a_i) = \![cosSim(embed(e_i),
       \hfill embed(a_i)) \geq 0.9]\! &\\
    &\text{DateMatch}(e_i,a_i) = \text{ExactMatch}(normDate(e_i),normDate(a_i))&
\end{flalign*}
Thus, the hybrid metric encompasses all previous challenges:
\begin{flalign*}
    &\text{evaluate}(e_i, a_i) = \begin{aligned}[t]
        & \text{ExactMatch}(e_i, a_i)
        \lor ((\text{NormExactMatch}(e_i, a_i)
    \end{aligned} &\\
    &\lor \begin{aligned}[t]
        & \text{FuzzyExactMatch}(e_i, a_i) 
        \lor \text{EmbedSimilarity}(e_i, a_i)) &\wedge &\\&\text{DateMatch}(e_i, a_i))
    \end{aligned} &
\end{flalign*}

\noindent where [] stands for the Inverson bracket, $norm$ transforms text into a standard format (e.g., removing punctuation, lowercasing), $fuzzyMatch$ computes a similarity score between answers based on edit distance, $cosSim$ stands for cosine similarity, $embed$ retrieves an embedding representation from a pre-trained sentence transformer\footnote{https://sbert.net/}, and $normDate$ converts dates into a standard format for comparison.

\subsection{Prompting}
\label{seq: prompting}
Large Language Models (LLMs) use the concept of prompting to tailor responses according to specific formats and requirements. This involves providing the model with structured input that guides it to produce desired outputs. By illustrating the task, expected behavior, and desired answer format through a few input-output examples, LLMs can excel in various straightforward question-answering tasks \cite{brown2020language}. In our case, the desired response is a chain-of-thought explanation and a few exact words from the text, which we call keywords, and are considered crucial words by the model. 
To achieve this, we define a structured dialogue framework process for interacting with the model, as:
    \begin{itemize}
        \item \textbf{System Message:} "To answer the given question, first generate a thought that explains the answer according to the text, then identify the most important words (keywords) from the text that helped you with your thought, and finally provide a short answer."
        \item \textbf{User Message:} "The following text might be useful in answering the question: [context] Question: [query]"
        \item \textbf{Assistant Message:} "Thought: [thought process], Keywords: [keywords], Short answer: [answer]"
    \end{itemize}

\subsection{Explainability Algorithm}
\label{algorithm}
Our algorithm extends the Leave One Out (LOO) method \cite{paes2024multi}, a powerful baseline in previous work (\cite{paes2024multi}). It follows a two-step process: first, identifying the \textit{sufficient regions} within the context, and second, detecting \textit{necessary keywords} in these regions. A key advantage of our approach is its constant complexity concerning the number of samples and model queries, which significantly reduces costs when using proprietary models.

\subsubsection{Sufficient Regions algorithm}
In order to identify the \textit{sufficient regions}, \(SR_i\), of the context we split it into p equal parts which are \textit{candidate regions}, \(CR_i\). We treat p as a hyperparameter (we selected p=3 for our experiments). Then we generate an answer for each candidate region \(s \in CR_i\) and if it
% : \(r_i = M(q_i, s), \forall s \in CR_i\). If an answer \(sa_i\)
is \textit{correct} then the corresponding region \(s\) is considered sufficient and is then added to the Sufficient Regions set: \(SR_i\). The details of the algorithm can be seen in Algorithm \ref{alg:sr}.

\begin{algorithm}
\caption{Sufficient Regions}
\label{alg:sr}
\begin{algorithmic}[1]
\State $\text{const } p \gets 3$
\Procedure{SR}{$q,c,e$} 
   \State $CR\gets c.split(p)$ \Comment{Candidate Regions} 
   \State $SR \gets set()$ 
   \For {s in $CR$}
   \State a = $M(q,s)$
   \If {$evaluate(e,a)==True$}
    \State $SR.add(s)$
   \EndIf
   \EndFor
   \State \textbf{return} $SR$
\EndProcedure
\end{algorithmic}
\end{algorithm}

\subsubsection{Necessary Keywords algorithm}
The \textit{sufficient regions} are usually sentence-level explanations (depending on the length of the input). To detect phrase-level explanations we adopt a slightly modified version of LOO and we call the results \textit{necessary keywords}. We apply this algorithm on every \textit{sufficient region} \(s\) of \(SR_i\). Instead of masking a single word or a predetermined number of words as in traditional LOO approaches, we mask q groups with equal number of words each (q=5 in our experiments), as in \textit{Sufficient Regions}. This way, the total number of API calls to the LLM is going to be $1+pxq$ (p=3 for SR, q=5 for NK, and one call to get the self-explanation keywords, 16 in total for our experiments), \textbf{independently} of the length of the input. For each group of words, we replace it with an underscore `\_', a process we call masking. We use the masked region \(s\) to generate an answer and if it is \textit{wrong} we add this group to the Necessary Keywords set: \(NK(s)_i\). The details of the algorithm can be seen in Algorithm \ref{alg:nk}.

\begin{algorithm}
\caption{Necessary Keywords}
\label{alg:nk}
\begin{algorithmic}[1]
\State $\text{const } q \gets 5$
\State $s \in SR$
\State $CK\gets s.split(q)$   \Comment{Candidate Keywords} 
\State $NK \gets set()$
\Procedure{NK}{$q,s,e$}
   \For {v in $CK$} 
   \State a = $M(q,s.mask(\text{v}))$
   \If {$evaluate(e,a)==False$}
    \State $NK.add(\text{v})$
   \EndIf
   \EndFor
   \State \textbf{return} $NK$
\EndProcedure
\end{algorithmic}
\end{algorithm}

%\vspace{-13pt}

\subsection{Faithfulness Evaluation}
To quantify the faithfulness of the response of the model, we compare the keywords \(K\) provided by the model with the \textit{sufficient regions} \(SR_i\) and the \textit{necessary keywords} \(NK_j\) identified with our proposed explainability algorithm \ref{algorithm}. We define the faithfulness score \(f_i\) for each question-context-answer triplet as follows:

%\vspace{-0.6em}
\[
f_i = \max_{s \in SR_{c_i}} \left\{ \frac{f_{SR}(s) + f_{NK}(s)}{2}\right\}
\]
\noindent where \(f_{SR}(s)\), \(f_{NK}(s)\) is the faithfulness score based on the \textit{sufficient regions} and the \textit{necessary keywords} respectively, defined as:

%\vspace{-1.2em}
\[f_{SR}(s) = \{ 1 \mid \exists k \in K \text{ such that } k \subseteq s, \text{ otherwise } 0 \}
\]

%\vspace{-1.4em}
\[
f_{NK}(s) = \frac{1}{|NK_s|} \sum_{t \in {NK_s}} g(t)
\]

\noindent where
\(
g(t) = \{1 \mid \exists k \in K \text{ such that } k \subseteq t, \text{ otherwise } 0\}
\).

\noindent The overall faithfulness score \(F\) for the dataset \(\mathcal{D}\) is then given by the average faithfulness score over all question-context pairs:

\[
F = \frac{1}{N} \sum_{i=1}^N f_i
\]

% \noindent This overall faithfulness score \(F\) provides a measure of how well the model's responses align with the sufficient regions and necessary keywords identified by our explainability algorithm across the entire dataset.

\noindent This overall faithfulness score \(F\) measures how well the responses of the model align with the sufficient regions and necessary keywords identified by our explainability algorithm across the entire dataset.

% \subsection{Visualization}
% We color the \textit{sufficient regions} in green, the \textit{necessary keywords} in blue, and highlight the self-generated keywords of the model in bold. As shown in Figure \ref{fig: example_thought}, this method visually represents how the self-explanations of the model align or not with the important context regions, along with the faithfulness score.

\subsection{Visualization}
We color \textit{sufficient regions} in green, \textit{necessary keywords} in blue, and highlight model-generated keywords in bold. Figure \ref{fig: example_thought} illustrates this method, showing how the self-explanations of the model align or not with the important context regions, along with the faithfulness score.

\section{Experiments}
\label{seq: experiments}

We have performed preliminary experiments which are still in progress. The OpenAI API \footnote{https://platform.openai.com/docs/api-reference/} was used to implement this framework, configuring interactions based on 
\ref{seq: prompting}. Currently, we have used 790 samples, only 311 of them are in the \textit{Retrieval-Hard subset} \ref{seq: retrievalhardsubset} (GPT-4o failed to answer without external context). We evaluate GPT-3.5 on the same subset.
We produced explanations with our multi-step approach and the success rate for each step can be seen in Table \ref{tab:results}.
An explanation might fail for three reasons: the model answered incorrectly even when the original context was given to it (row 2), no \textit{sufficient regions} were identified (row 3), or no \textit{necessary keywords} were found (row 4).
% An explanation might fail due to one of three reasons: the model responded incorrectly even when the original context was given to it (row 2), the set of \textit{sufficient regions} was empty (row 3), or the set of \textit{necessary keywords} was empty (row 4).
Our proposed explainability method is successful 100 out of 224 times and 149 out of 227 times respectively for GPT-3.5 and GPT-4o. There is a trade-off between API-calls/cost and explainability success. Our hyperparameter ($p=3, q=5$) results in 16 API calls per sample, achieving a 45\% explainability success rate on average. This rate can be improved by adjusting $p, q$.
For a fair comparison, we evaluate the faithfulness of GPT-3.5 and GPT-4o on the common subset of successfully explained samples only.
The total common successful samples are 62.
Preliminary results indicate that GPT-4o shows higher faithfulness than GPT-3.5, aligning more accurately with key context regions and keywords. This suggests advancements in model training and algorithmic refinement in newer LLM versions.

\begin{table}[ht]
\centering
\begin{tabular}{@{}lcc@{}}
\toprule
\textbf{Models} & \textbf{GPT-3.5} & \textbf{GPT-4o} \\
\midrule
Retrieval-Hard Subset & 311 & 311 \\
\midrule
Successful samples (original context) & 224 & 227 \\
\midrule
Successful Sufficient Regions & 119 & 177 \\
\midrule
Successful Necessary Keywords & 100 & 149 \\
\midrule
Common successful samples & 62 & 62 \\
\midrule
Faithfulness score (common) & 0.653 & 0.691 \\
\bottomrule
\end{tabular}
\caption{Success rate of each stage of the faithfulness evaluation process and faithfulness score on the common subset.}
\label{tab:results}
\end{table}

%\vspace{-25pt}
\balance

\section{Related Work}
\label{seq: relatedwork}
Our
focus is on local post-hoc explanations for LLMs, given that textual output is the sole result.
LLMs can provide themselves explanations in line with subsequent outputs, referred to as chain-of-though \cite{wei2022chain}, and can perform in-context few-shot learning by using prompts, where users illustrate the task using a few input-output examples with great success in various straightforward QA tasks \cite{brown2020language}.

Mosca et al. \cite{mosca2022shap} highlight the scarcity of studies on perturbation-based explanations for text inputs.
Ribeiro et al. \cite{ribeiro2016should} generate perturbed variations of the context to train an interpretable model that mimics the local predictions of the black-box model. 
% In NLP tass, the traditional Shapley value approach oversimplifies the impact of individual features, ignoring strong interactions among words that heavily depend on context. Instead of treating single words in isolation, it is essential to extend relevance assessments to multi-level tokens or entire sentences \cite{mosca2022shap}.
% Chen et al. \cite{chen2018shapley} proposed L-Shapley and C-Shapley to provide efficient model interpretation methods for NLP tasks. L-Shapley focuses on local interactions by perturbing neighboring features, while C-Shapley considers multi-level tokens and entire sentences.
In NLP tasks, traditional Shapley values oversimplify feature impact by ignoring contextual interactions among words. Instead, relevance assessments should extend to multi-level tokens or entire sentences \cite{mosca2022shap}. Chen et al. \cite{chen2018shapley} introduced L-Shapley and C-Shapley for better interpretation: L-Shapley examines local interactions through neighboring feature perturbation, while C-Shapley assesses multi-level tokens and full sentences.

Hierarchical Explanation via Divisive Generation (HEDGE) \cite{chen2020generating} exemplifies a SHAP-based method addressing the challenge of lengthy texts. HEDGE sequentially breaks down text into shorter phrases and words based on their weakest interactions, assigning relevance scores at each level to achieve a hierarchical explanation. Similarly, PartitionSHAP \footnote{https://github.com/slundberg/shap}, adopts a comparable approach by forming hierarchical coalitions of features and evaluating their interactions. Also, CaptumLIME, a modified version of LIME \cite{ribeiro2016should} adapted for text generation tasks using features from the Captum library \cite{miglani2023using}, allows users to define units for attribution within the input manually. It addresses sequence outputs by computing log probabilities for tokens in the output and summing them.

Paes et al. \cite{paes2024multi} extend perturbation-based methods to handle text outputs and long inputs, using scalarizers to map text outputs to real numbers for input importance assessments.
They use user studies and BARTScore \cite{yuan2021bartscore} 
to measure the likelihood of a reference text conditioned on the generated text \cite{yuan2021bartscore}.
Ribeiro et al. \cite{ribeiro2016should} examine faithfulness by comparing the features the model claims to rely on with those identified by an explanation technique, using a restricted set of \textit{gold features}. Schnake et al. \cite{pmlr-v162-ali22a} observe that a higher area under the activation curve indicates more faithful explanations.

\section{Summary and Future Work}

\label{seq: conclusion_futurework}
We introduce a novel approach for assessing LLM faithfulness using local and self-explanations. Inspired by the leave-one-out technique, our approach identifies essential and sufficient parts of the context affecting model answers. We propose a metric for evaluating faithfulness by comparing these parts with the model's self-explanations. Experiments with the Natural Questions dataset demonstrate the approach's effectiveness. Future work will involve testing on broader QA datasets \cite{berant2013semantic, joshi2017triviaqa, 10.1007/978-3-319-24027-5_20}, analyzing the trade-off between explanation success rate and API calls, and comparing our method with existing baselines.

\begin{acks}
This work has been partially supported by project MIS 5154714 of the National Recovery and Resilience Plan Greece 2.0 funded by the European Union under the NextGenerationEU Program.
\end{acks}

\bibliographystyle{ACM-Reference-Format}
\bibliography{sample-base}

% \appendix









\end{document}